%% file: main.tex
\pdfoutput=1
\documentclass[letterpaper, 10 pt, conference]{ieeeconf}  

\IEEEoverridecommandlockouts                              
\usepackage{graphicx}
\usepackage[font=footnotesize]{caption}
\usepackage{subcaption}
\usepackage{float}
\usepackage{amssymb}
\usepackage{mathtools}
\usepackage{cite}
\usepackage{xcolor}
\usepackage{graphicx}
\usepackage[utf8]{inputenc} 
\usepackage[T1]{fontenc}
\usepackage{makecell}

\newcommand{\pointcloud}{C}
\newcommand{\transx}{x}
\newcommand{\transy}{y}
\newcommand{\orient}{\theta}
\newcommand{\mappingfunction}{f}
\newcommand{\gtdistancethreshold}{\rho}
\newcommand{\prvec}{v}
\newcommand{\orientvec}{w}
\newcommand{\orientvecanchor}{\orientvec_A}
\newcommand{\orientvecsame}{\orientvec_S}
\newcommand{\prvecanchor}{\prvec_A}
\newcommand{\prvecsame}{\prvec_S}
\newcommand{\prvecdifferent}{\prvec_D}
\newcommand{\distsame}{\delta_S}
\newcommand{\distdiff}{\delta_D}
\newcommand{\yawdiff}{\delta\theta}
\newcommand{\prloss}{L_{pr}}

\newcommand{\orientloss}{L_{\theta}}

\newcommand{\prnet}{N_{DE}}

\newcommand{\imganchor}{I_A}
\newcommand{\imgsame}{I_S}
\newcommand{\imgdifferent}{I_D}
\newcommand{\oreos}{OREOS }
\newcommand{\oreosa}{OREOS}

\newcommand{\sevensensesymbol}{2}
\newcommand{\aslsymbol}{1}
\newcommand{\zenith}{\theta}
\newcommand{\azimuth}{\varphi}
\title{\LARGE \bf
OREOS: Oriented Recognition of 3D Point Clouds in Outdoor Scenarios
}

\author{Lukas Schaupp$^{\aslsymbol}$, Mathias B\"{u}rki$^{\aslsymbol,\sevensensesymbol}$, Renaud Dub\'e$^{\aslsymbol, \sevensensesymbol}$, Roland Siegwart$^{\aslsymbol}$, Cesar Cadena$^{\aslsymbol}$
\thanks{This research has received funding from the EU H2020 research project under grant agreement No 688652, the Swiss State Secretariat for Education, Research and Innovation (SERI) No 15.0284}
\thanks{$^{\aslsymbol}$Autonomous Systems Lab (ASL), ETH Z\"{u}rich, Switzerland {\tt\footnotesize \{firstname.lastname\}@ethz.ch}}
\thanks{$^{\sevensensesymbol}$Sevensense Robotics AG, Switzerland {\tt\footnotesize \{firstname.lastname\}@sevensense.ch}}
}

\begin{document}

\maketitle
\thispagestyle{empty}
\pagestyle{empty}

\begin{abstract}
We introduce a novel method for oriented place recognition with 3D LiDAR scans.
A Convolutional Neural Network is trained to extract compact descriptors from single 3D LiDAR scans. These can be used both to retrieve near-by place candidates from a map, and to estimate the yaw discrepancy needed for bootstrapping local registration methods.
We employ a triplet loss function for training and use a hard-negative mining strategy to further increase the performance of our descriptor extractor.
In an evaluation on the NCLT and KITTI datasets, we demonstrate that our method outperforms related state-of-the-art approaches based on both data-driven and handcrafted data representation in challenging long-term outdoor conditions. 
\end{abstract}

\input{introduction}
\input{related_work}
\input{theory}
\input{experiments}
\input{conclusions}





\bibliographystyle{IEEEtran}
\input{main.bbl}
\end{document}

%% file: introduction.tex
\section{Introduction}
\label{sec:intro}

%
%

Global localization constitutes a pivotal component for many autonomous mobile robotics applications.
It is a requirement for bootstrapping local localization algorithms and for re-localizing robots after temporarily leaving the mapped area.
Global localization can furthermore be used for mitigating pose estimation drift through loop-closure detection and for merging mapping data collected during different sessions.
Prior-free localization is especially challenging for autonomous vehicles in urban environments, as GNSS-based localization systems fail to provide reliable and precise localization near buildings due to multi-path effects, or in tunnels or parking garages due to a lack of satellite signal reception.
%
%
Due to their rich and descriptive information content, camera images have been of great interest for place recognition, with mature and efficient data representations and feature descriptors evolving in recent years.
However, visual place recognition algorithms struggle to cope with strong appearance changes that commonly occur during long-term applications in outdoor environments, and fail under certain ill-lighted conditions \cite{Lowry:2016:VPR:3046690.3046765}.
In contrast to that, active sensing modalities, such as LiDAR sensors, are mainly unaffected by appearance change ~\cite{DBLP:conf/icra/McManusFB11}. 
Efficient and descriptive data representations for place recognition using LiDAR point clouds remain, however, an open research question~\cite{zhou2018voxelnet, bosse13,DBLP:journals/corr/abs-1804-09557}.
%
%
In contrast to our work, typical place recognition methods do not always explicitly deal with the full problem of estimating a 3 DoF transformation ~\cite{YiWa17} ~\cite{DBLP:journals/corr/abs-1804-03492} ~\cite{20.500.11850/302166}.

This paper addresses the aforementioned issue by presenting a data-driven descriptor for sparse 3D LiDAR point clouds which allows for long-term 3 DoF metric global localization in outdoor environments.
Specifically, our method allows us to estimate the relative orientation between scans. 
\begin{figure}
\centering
\includegraphics[width=0.45\textwidth]{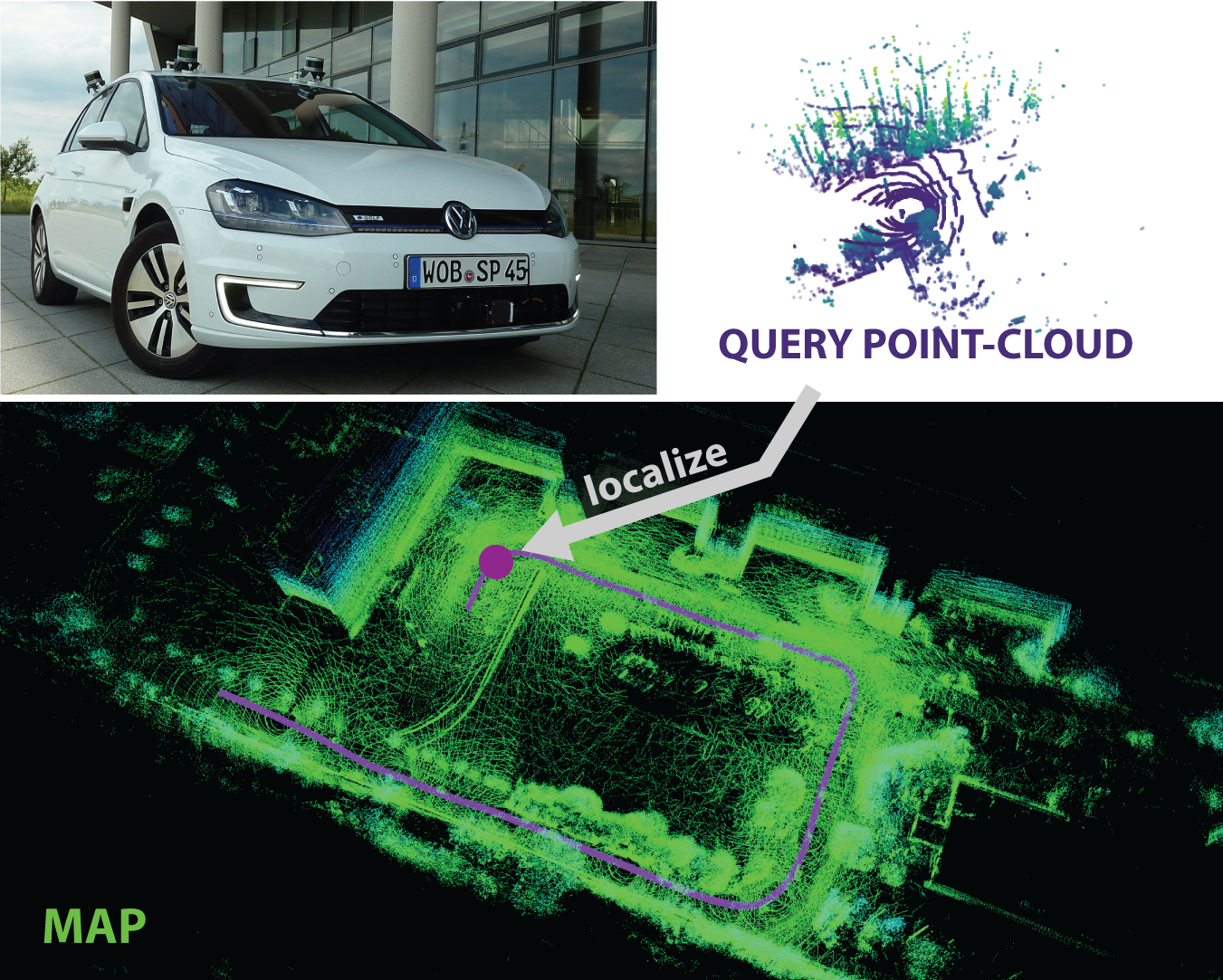}
\caption{
We aim at accurately localizing our vehicle in a map build from previously collected LiDAR point clouds. %
A query point cloud scan is fed through the \oreos pipeline, yielding a compact descriptor that allows to retrieve near-by place candidates from the map, and estimate the yaw angle discrepancy.
With this information, a local registration method, such as ICP, can be bootstrapped and used for subsequent high accuracy localization along the traversal.
\label{fig:scenario_front}}
\vspace{-5mm}
\end{figure}
Our novel data-driven metric global localization  descriptor is fast to compute and robust with respect to long-term appearance changes of the environment, and shows similar place recognition performance compared to other state-of-the-art LiDAR place recognition approaches. 
Additionally our architecture provides orientation descriptors capable of predicting a yaw angle estimation between two point clouds realizations of the same place.
Our contributions can be summarized as follows:
\begin{itemize}
\item We present \oreosa: an efficient data-driven architecture for extracting a point cloud descriptor that can be used both for place recognition purposes and for regressing the relative orientation between point clouds. 
%
\item In an evaluation using two public dataset collections, we demonstrate the capability of our approach to reliably localize in challenging outdoor environments across seasonal and weather changes over the course of more than a year. We show that our approach works even under strong point cloud misalignment, allowing the arbitrary positioning of a robot.
\item A computational performance analysis showing that our proposed algorithm exhibits real-time capabilities and performs similarly to other state-of-the-art approaches in place recognition performance while providing robustness and better performance in the metric global localization.
\end{itemize}
The paper is structured as follows.
After an overview over related work, we describe the \oreos metric global localization pipeline in Section~\ref{sec:meth}, before presenting our evaluation results in Section~\ref{sec:exp}. 

%% file: related_work.tex
\section{related work}
\label{sec:related_work}
We subdivide the related work into two categories.
First we discuss related approaches for solving the place recognition problem. 
Then we show related work on pose estimation for 3D point clouds. 

\subsubsection*{Place Recognition}
Early approaches to solving the place recognition problem with LiDAR data have, analogous to visual place recognition, focused on extracting keypoints on point clouds and describing their neighborhood with structural descriptors~\cite{KSH12}.
Along this vein, Bosse~\textit{et al.} used a 3D Gestalt descriptor~\cite{BZ13}, Steder~\textit{et al.}~\cite{St+10} and Zhuang~\textit{et al.}~\cite{CZZW18} transformed a point cloud to a range or bearing-angle based image by extracting local-invariant ORB~\cite{Rublee:2011:OEA:2355573.2356268} features for database matching.
The strength of these approaches is the explicit extraction of low-dimensional data representations that can efficiently be queried in a nearest neighbor search.
The representation, however, is handcrafted, and may thus not capture all relevant information efficiently in every application scenario.
Furthermore, the dependence on good repeatability inherent to the keypoint detection constitutes an additional challenge for these approaches, especially if the sensor viewpoint is slightly varying. The drawbacks of keypoint-based approaches can be tackled by employing a segmentation of the point clouds, and computing place dependent data representations on these segments for subsequent place recognition~\cite{Man09,DBLP:journals/ral/DubeGSGSCN18,DBLP:journals/corr/abs-1804-09557, DBLP:journals/corr/DubeDSNSC16}.
As a requirement for a proper segmentation, giving the sparsity of the data, these methods require the subsequent point clouds to be temporarily integrated and smoothed.
In contrast to that, our data representation for place recognition can be computed directly from a single point cloud scan, which obviates any assumption on how the LiDAR data is collected and processed, and even allows to obtain localization without movement.
Related approaches that compute handcrafted global descriptors for place recognition from aggregated point clouds are presented by Cop~\textit{et al.}~\cite{20.500.11850/302166}, who generate distributed histograms of the intensity channel.
Along a similar vein, Rohling~\textit{et al.}~\cite{Ro+15} represent each point cloud with a global 1D histogram, and Magnusson~\textit{et al.}~\cite{Man09} use the transform-based surface feature NDT (Normal Distribution Transformation). 
Further  global  descriptors  such as  GASD~\cite{DT17}, and the extended FPFH - VFH~\cite{Ru+10} can be also used for the task of place recognition.
Recent advances in machine learning have opened up new possibilities to deal with the weaknesses of handcrafted data presentation for place recognition with LiDAR data.
Employing a DNN (Deep Neural Network) to learn a suitable data representation from point clouds for place recognition allows for implicitly encoding and exploiting the most relevant cues in the input data. 
Within this field of research Yin~\textit{et al.} LocNet~\cite{YiWa17} use semi-handcrafted range histogram features as an input to a 2D CNN (Convolutional Neural Network), while Uy~\textit{et al.} use a NetVLAD~\cite{Arandjelovic16} layer on top of the PointNet~\cite{GGG16} architecture ~\cite{DBLP:journals/corr/abs-1804-03492}. 
Furthermore, Kim~\textit{et al.}~\cite{gkim-2019-ral} recently presented the idea to transform point clouds into scan context images~\cite{scan_context} and feed them into a CNN for sovling the place recognition problem.
Apart from the work by Uy~\textit{et al.}, all these approaches depend on a precomputed handcrafted descriptor, which may not represent all relevant information in an optimal, in this case most compact, manner.
In contrast to that, we refrain from any pre-processing of the point clouds and directly employ our DNN on the raw 2D projected LiDAR data.
In comparison to LocNet, our data-driven method is capable of learning a descriptor that is used both for fetching a nearest neighbour place and for estimating the orientation, which is not possible after the computation of the inherently rotation invariant histogram representation.
\subsubsection*{Pose Estimation}
Common approaches to retrieve a 3 DoF pose from LiDAR data employ either local features extraction such as FPFH~\cite{Rusu2009} and feature matching using RANSAC~\cite{Fischler:1981:RSC:358669.358692}, or use handcrafted rotation variant global features such as VFH~\cite{Ru+10} or GASD~\cite{DT17}. 
An overview of recent research on 3D pose estimation and recognition is given by Han~\textit{et al.}~\cite{DBLP:journals/corr/abs-1802-02297}.
Velas~\textit{et al.} \cite{velas2018cnn} propose to use a CNN to estimate both translation and rotation between successive LiDAR scans for local motion estimation.
In contrast to this, we aim for solving the metric global localization problem, and demonstrate that the best performance is obtained by a combination of learning and classical registration approaches.


%% file: theory.tex
\section{Methodology}
\label{sec:meth}
We first define the problem addressed in this paper, and outline the pipeline we propose for solving it, before elaborating in detail on our Neural Network architecture and the training process.
%

\subsection{Problem Formulation}
\label{sec:methodology:problem_formulation}
Our aim is to develop a metric global localization algorithm, yielding a 3 DoF ($\transx, \transy, \orient$) in the map reference frame from a single 3D LiDAR point cloud scan $\pointcloud$. 
%
%
This can formally be expressed with a function $\mappingfunction$ as follows:
\begin{equation}
\transx, \transy, \orient \coloneqq \mappingfunction(\pointcloud) \textit{, with } \transx, \transy, \orient \in \mathbb{R}
\end{equation}
\begin{figure*}[htp]
\centering
\includegraphics[scale=0.6, trim = 0mm 10mm 0mm 0mm]{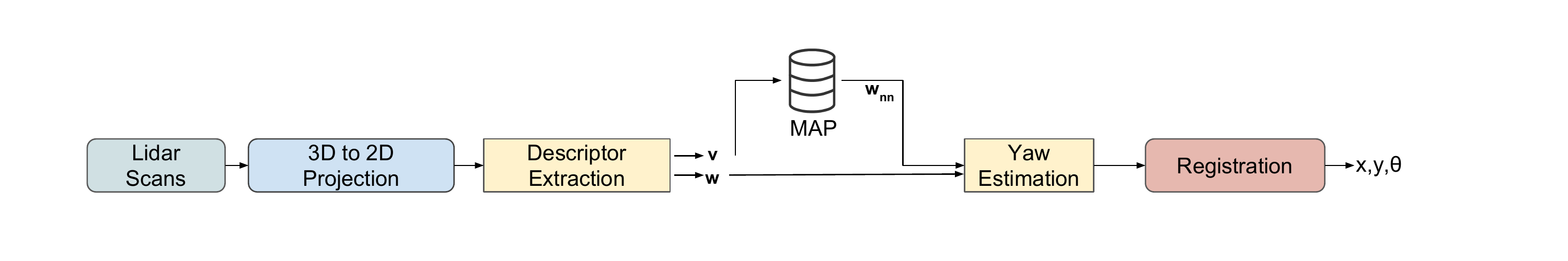}
\caption{In a first step, we project the current 3D LiDAR scan onto a 2D range image. 
In a second stage, a Convolutional Neural Network is leveraged for extracting two compact descriptors $\prvec$, and $\orientvec$. 
The former is used in a k-Nearest-Neighbor (kNN) search to retrieve the $\orientvec_{nn}$ from the closest place candidate in the map. 
Both descriptors $\orientvec$ and $\orientvec_{nn}$ are then fed to into a Yaw Estimation Neural Network to estimate the yaw angle discrepancy between the query point cloud and the point cloud of the nearest place in the map. 
Finally, an accurate 3 DoF pose estimate is obtained by applying a local registration method, using the orientation estimate and the planar $x$ and $y$ coordinates from the nearest map place as an initial guess.}
\label{fig:system_overview}
\vspace{-5mm}
\end{figure*}
To solve this problem, we divide function $\mappingfunction$, as depicted in Figure~\ref{fig:system_overview}, into the following four sequential components: a) Point Cloud Projection b) Descriptor Extraction c) Yaw Estimation, and d) Local Point Cloud Registration. 
The Point Cloud Projection module converts the input LiDAR point cloud scan $\pointcloud$, given by a list of point coordinates $p_x$, $p_z$ and $p_z$ within the sensor frame, onto a 2D range image using a spherical projection model:
\begin{equation} 
\varphi = atan (\frac{p_y}{p_x})
\end{equation}
\begin{equation} 
\rho = asin (\frac{y_s}{\sqrt{p_x^2+p_y^2+p_z^2}})
\end{equation}
The zenith $\zenith$ and azimuth $\azimuth$ angles are directly mapped onto the image plane, yielding a 2D range image.
%
For our work we use the whole 360 degree field of view given by one point cloud scan and the range information of the sensor. 

The Descriptor Extraction module aims at deriving a compact representation for place-, and orientation related information from the input data.
This is achieved by employing a Convolutional Neural Network, taking the normalized 2D range image as an input and generating two compact descriptors vectors $\prvec$, and $\orientvec$ respectively. 
While $\prvec$ represents rotation invariant place dependent information, $\orientvec$ encodes rotation variant information used for determining a yaw angle discrepancy in a latter stage of the pipeline.

The place specific vector $\prvec$ can be used to query our map for nearby place candidates, yielding a map position $\transx_{nn}$, $\transy_{nn}$, and orientation descriptor $\orientvec_{nn}$ of a nearest neighbor place candidate.
In the subsequent step, the Yaw Estimation module estimates a yaw angle discrepancy $\yawdiff$ between the query point cloud $\pointcloud$, and the point cloud associated with the retrieved nearest place in the map.
For this, the two orientation descriptors $\orientvec$, and $\orientvec_{nn}$ are fed into a small, fully connected Neural Network, which directly regresses a value for $\yawdiff$.

The position of the map place candidate $\transx_{nn}$, and $\transy_{nn}$, together with the yaw discrepancy $\yawdiff$, can then be used as an initial condition for further refining the pose estimation, yielding the desired highly accurate 3DoF pose estimate $\transx$, $\transy$, and $\orient$ of the point cloud $\pointcloud$ in the map coordinate frame.
%

Note that in our map, the place dependent descriptors $\prvec$ extracted from point cloud scans of a map dataset are organized in a kd-tree for fast nearest neighbor search.
Retrieving the orientation descriptor $\orientvec_{nn}$ of a map place candidate can be achieved by a simple look-up table.
\subsection{Network Architecture}
\label{sec:methodology:dnn}
The network architecture of the CNN used for the descriptor extraction is based on the principles described in~\cite{Si+15, AC17}.
We use a combination of 2D Convolutional and Max Pooling Layers for feature extraction.
Subsequent fully connected layers map the features into a compact descriptor representation as depicted~Figure~\ref{fig:network_architecture}. 
%
%
As proposed by Simonyan~\textit{et al.}~\cite{Si+15}, we use smaller filters rather than larger filters as well as designed the network around the receptive field size.
Additionally, we use asymmetric pooling layers at the beginning of the architecture to further increase the descriptor retrieval performance.

In contrast to that, our Yaw Estimation network is composed of two fully-connected layers.
%
%


\subsection{Training the \oreos descriptor}
\label{sec:meth:training}
The two neural networks pursue two orthogonal goals, namely finding a compact place dependent descriptor representation for $\prvec$, and finding a compact orientation dependent descriptor representation for $\orientvec$.
For each of these two goals, a loss term is defined, denoted by the place-recognition loss $\prloss$, and orientation loss $\orientloss$, respectively.
\subsubsection*{Place-Recognition Loss}
\begin{figure*}[htp]
\centering
\includegraphics[scale=0.55, trim = 40mm 40mm 0mm 30mm]{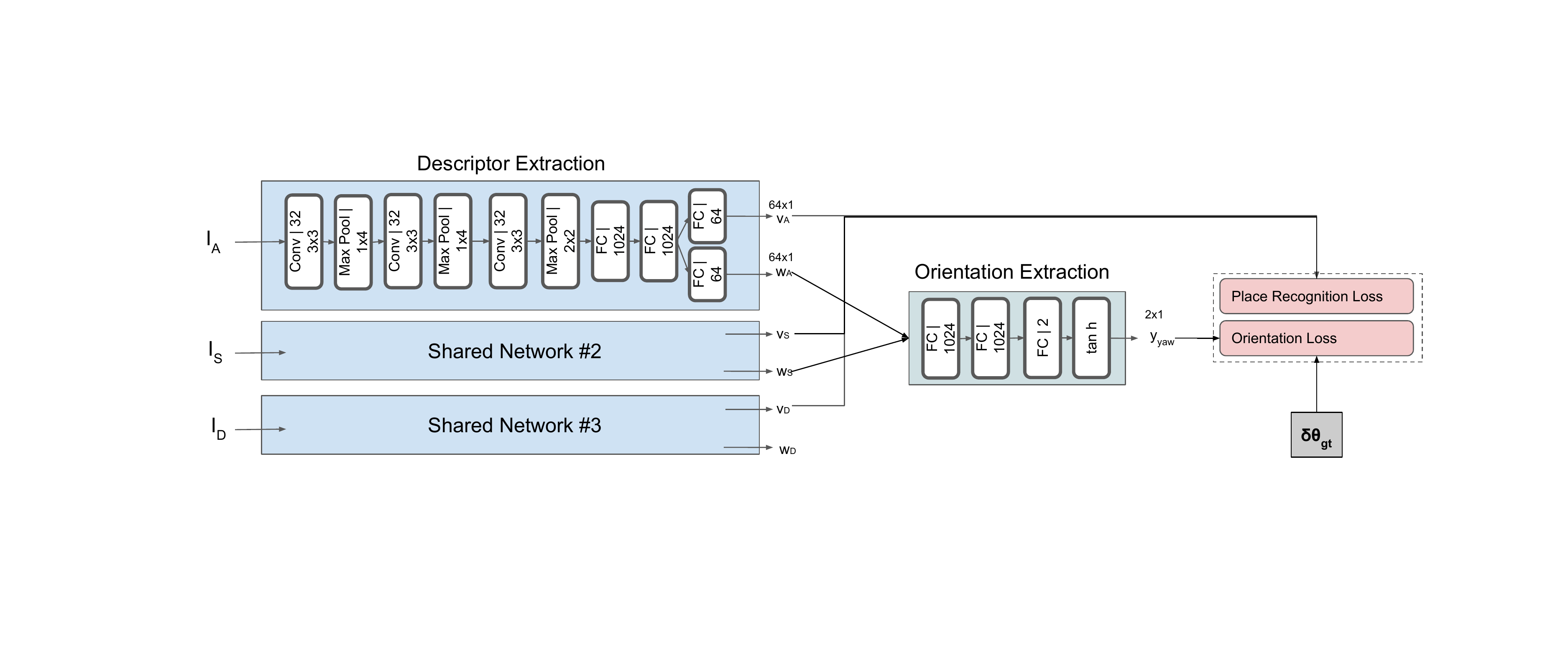}
\caption{
Our proposed network architecture for descriptor extraction is composed of three convolutional and two fully connected layers.
The projected 2D range images ($\imganchor, \imgsame, \imgdifferent$) representing an anchor, a similar and a dissimilar point cloud sample, are compressed into a descriptor of dimension 2 x 64 x 1 which can be used for global localization and orientation estimation.
The three place recognition vectors $\prvecanchor, \prvecsame, \prvecdifferent$ are used to compute the Place Recognition Loss $\prloss$, while the two orientation vectors from the point clouds from similar places, that is $\orientvecanchor$ and $\orientvecsame$, are fed into the Orientation Extraction network.
The latter estimates a yaw angle discrepancy $\yawdiff$, which is used to compute the Orientation Loss and compared with the yaw angle discrepancy ground truth label $\yawdiff_{gt}$.
%
%
We abbreviated all layers of our network, where FC represents a fully connected, and Conv represents a convolutional layer. %
Note that PreLU activation functions are used unless otherwise stated.}
\label{fig:network_architecture}
\end{figure*}

To train our network for the task of place recognition, we use the triplet loss method ~\cite{Hoffer2015DeepML}.
The loss-function is designed to steer the network towards pushing similar and dissimilar point-cloud pairs close together and far apart in the resulting vector space. 
Let $\prnet$ denote our descriptor extraction network, and let $\imganchor$ denote an anchor range image, $\imgsame$ a range image from a similar place, and $\imgdifferent$ a range image from a dissimilar place. 
The Neural Network $\prnet$ transforms these input images into three place dependent output descriptors $\prvecanchor, \prvecsame, \prvecdifferent$ as depicted in Figure~\ref{fig:network_architecture}.
We further define $\distsame$ as the euclidean distance between descriptors of the anchor and similar place, and $\distdiff$ as the distance between descriptors of the anchor and the dissimilar one, and $m$ as a margin parameter for separating similar and dissimilar pairs.
%
%
%
%
%
The triplet loss can then be defined as follows:
\begin{equation}
    \begin{array}{r c l}
       &L_{pr}(D_p,D_n) = D_p^2 - D_n^2 + m\\[5pt]
        &\mbox{\textit{with} } D_p=\| {f(I_i^A)-f(I_i^S)} \|_2^2\\[5pt]
        &\mbox{\textit{and} } D_n=\| {f(I_i^A)-f(I_i^D)} \|_2^2
    \end{array}
    \label{eq:1}
\end{equation}
\subsubsection*{Orientation Estimation Loss}
\label{sec:meth:training:orient_loss}
As we want to predict an orientation estimate, we are implementing a $L_\theta$ regression loss function. 
For this task, we add an additionally fully-connected layer at the end of the triplet network. In this case we make only use of the anchor image $I_A$ and the similar image $I_S$ and obtain our rotation dependent descriptors $w_{A}$ and $w_{S}$ from our descriptor extraction network $\prnet$. 
We then feed the obtained descriptors $w_{A}$ and $w_{S}$ into a additional orientation estimation network that yields the yaw angle discrepancy descriptor $y_{yaw}$ between both given point clouds which is then compared to our ground truth yaw discrepancy angle  $\yawdiff_{gt}$.
By transforming the ground truth yaw angle  $\yawdiff_{gt}$ into the euclidean space, the ambiguity between 0 and 360 degree angles is avoided, which would result in false corrections during training. 
The orientation loss term is defined as follows:
\begin{equation}
    \begin{array}{r c l}
L_{\theta}(y_{yaw}, \yawdiff_{gt}) = \frac{1}{2}((y_{yaw,0} - cos(\yawdiff_{gt}))^2 \\[5pt]
+ (y_{yaw,1} - sin(\yawdiff_{gt}))^2)
    \end{array}
\end{equation}
where $y_{yaw,i}$ represents the i-th index of our yaw angle discrepancy descriptor $y_{yaw}$.
%
\subsubsection*{Joint Training}
As it is the goal of our proposed metric localization algorithm to both achieve a high localization recall with an accurate yaw angle estimation, we learn the weights of both Neural Network architectures in a joint training process.
For this, both loss terms are combined into a joined loss $L$ as follows:
\begin{equation}
\label{eq:combined_loss}
    \begin{array}{r c l}
L = \prloss + \orientloss
    \end{array}
\end{equation}
The joint training consists of a three-tuple network, whereas we sample point clouds based on the euclidean distance of their associated ground truth poses and a predefined distance threshold $\gtdistancethreshold$. 
%
%
The three point clouds are fed after the 2D projection into the Descriptor Extractor network, and the corresponding three place dependent output vectors $\prvec_{A}$, $\prvec_{S}$, and $\prvec_{D}$ are fed into the Place-Recognition Loss $\prloss$.
In contrast to that, the two orientation specific vectors $\orientvec_{A}$, and $\orientvec_{S}$ from the two close-by point clouds are fed into the Orientation Loss $\orientloss$.
%
The combined loss $L$ is then evaluated as described in Equation~\ref{eq:combined_loss}.
We use ADAM~\cite{DBLP:journals/corr/KingmaB14} as a learning optimizer and use a learning rate of alpha = 0.001.
We convert our range data to 16 bit and normalize the channel before training. 
To achieve rotation invariance for our place recognition descriptor $v$ and generate training data for our yaw angle discrepancy descriptor $w$, we employ data augmentation by randomly rotating the input image around its yaw-axis.

%

%% file: experiments.tex
\section{experiments}
\label{sec:exp}
Our experimental evaluation pursues the following goals:
a) In a comparison of our proposed metric global localization algorithm with related state-of-the-art techniques, we demonstrate that our approach not only outperforms existing feature-based algorithms, but that it is also computationally less expensive.
%
b) In addition to that, we provide valuable insights of the place-recognition and orientation estimation performance by performing a separate in depth analysis of two core modules of our pipeline dedicated at deriving a compact place dependent, and a compact orientation dependent descriptor, respectively.
%

Before addressing these two evaluation foci in detail, a brief overview of the two dataset collections used and the respective sensor setups is provided.

\subsection{Dataset Collections}
\label{sec:exp:subsec:datasets}
We use the following two dataset collections for our experiments:
\subsubsection{KITTI}
\label{sec:exp:subsec:datasets:subsubsec:kitti}
The KITTI dataset collection contains recordings from several drivings through the urban areas of Karlsruhe ~\cite{Geiger2013IJRR}. 
The point clouds are recorded by a Velodyne 64 HDL sensor at 10 Hz, placed on the center of the car's roof. 
Ground truth poses are provided by a RTK GPS sensor.
\subsubsection{NCLT}
The University of Michigan North Campus Long-Term Vision and LIDAR Dataset ~\cite{CUE16} consists of 27 recordings collected by driving a Segway platform through the indoor and outdoor of the University campus over the course of 14 months.
A Velodyne HDL-32 sensor provides point clouds at 10 Hz, and ground truth trajectories are provided by a globally optimized SLAM solution fusing RTK GPS with co-registered LiDAR point clouds.

%

\subsection{Data Sampling for Training}
Training triplet network structures requires sampling three-tuples of anchor, similar, and dissimilar pairs, as described in Section~\ref{sec:meth:training}.
Two point clouds are considered similar, if their ground-truth poses are within $1.5m$.
In the first training stage, dissimilar point clouds are sampled randomly from outside the $1.5m$ radius around the anchor sample.
This is followed by a second training stage, where dissimilar point clouds are sampled from within a $2-5m$ radius around the anchor sample.
This hard-negative mining strategy is able to boost the network performance by training with three-tuples that are harder to distinguish in the later stage of convergence.
%
%
%
%
%
For the NCLT dataset collection, we train our model with data from a subarea of the campus using the 2012-01-08 and 2012-01-15 datasets. 
Validation has been done on 2012-12-01, while we use six different datasets (2012-01-22, 2012-02-04, 2012-03-25, 2012-03-31, 2012-10-28 and 2012-11-17) for our final evaluation. 
The campus subarea used in the six validation datasets is different from the area used for training.
Furthermore, we have downsampled the data, such that for each query point cloud, there is exactly one true-positive map point cloud, and any two query point clouds in the same dataset are at least 3 meters apart.
In case of the KITTI dataset collection, only Sequence 00 revisits the same places again, and can thus be used for proper localization evaluation. 
%
%
Sequences 03-08 are used for training, while Sequence 02 is used for validation.
For the evaluation, point clouds from the first $170s$ of Sequence 00 are used to generate the map, i.e., to populate the KD-tree.
The remaining point clouds are used for localization queries.
This split of Sequence 00 prevents any self-localization, as the vehicle starts to revisit previously traversed areas after $170s$.
Analogous to the NCLT datasets, the query point clouds are sampled to be at least $3m$ apart. 
\subsection{Baselines}
We compare our metric global localization algorithm with two versions of LocNet~\cite{YiWa17}:
\begin{itemize}
    \item \textbf{LocNet (base)}: feeds handcrafted rotation invariant histogram-based range images into a CNN. 
    We have reimplemented LocNet with the network architecture as described in Yin et al. for the base model of LocNet.
    \item \textbf{LocNet++}: We retrained the original LocNet model following our training procedure, i.e., by using the triplet loss and hard negative mining.
\end{itemize}
In contrast to our work, LocNet is only able to provide a nearest place candidate in a map, but no metric pose estimate.
An orientation estimate can, however, be generated using local handcrafted features together with RANSAC:
\par \begin{itemize}
    \item \textbf{FPFH + RANSAC} \cite{Li16}: we generate for each point a local feature and use RANSAC to obtain the prior pose estimate from the inlier set.
\end{itemize}
Both (FPFH and RANSAC) are implemented using the PCL library ~\cite{Ru+11}, while LocNet's histogram generation is implemented using Matlab. 
Pose estimates generated by our metric global localization algorithm, and by LocNet in combination with FPFH and RANSAC, are further refined with point-to-plane ICP, yielding accurate 3 DoF pose estimates.

\begin{figure}[!tbp]
\centering
  \begin{subfigure}[b]{0.4\textwidth}
\includegraphics[width=\textwidth]{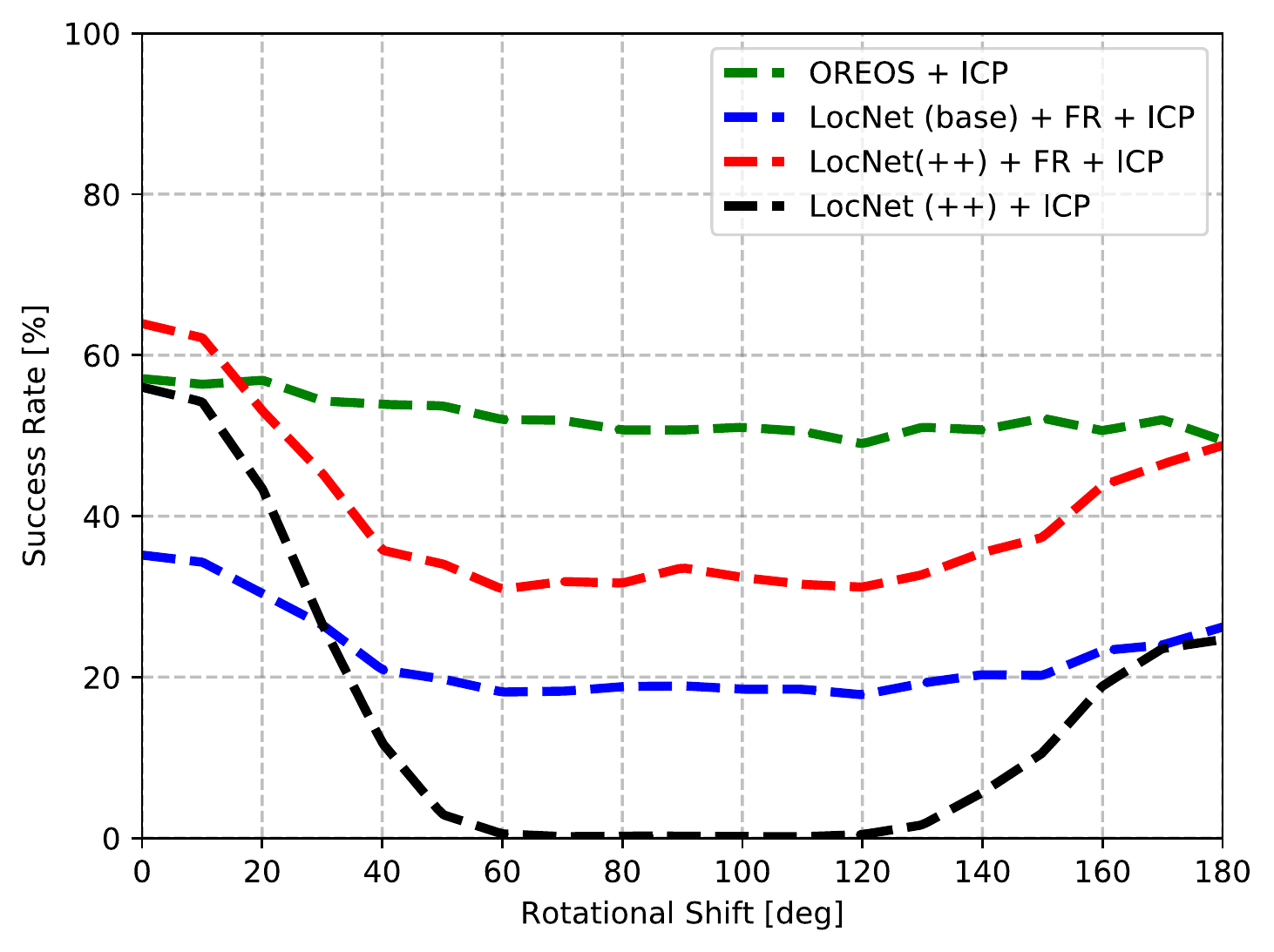}
    \label{fig:f1}
  \end{subfigure}
  \begin{subfigure}[b]{0.4\textwidth}
\includegraphics[width=\textwidth]{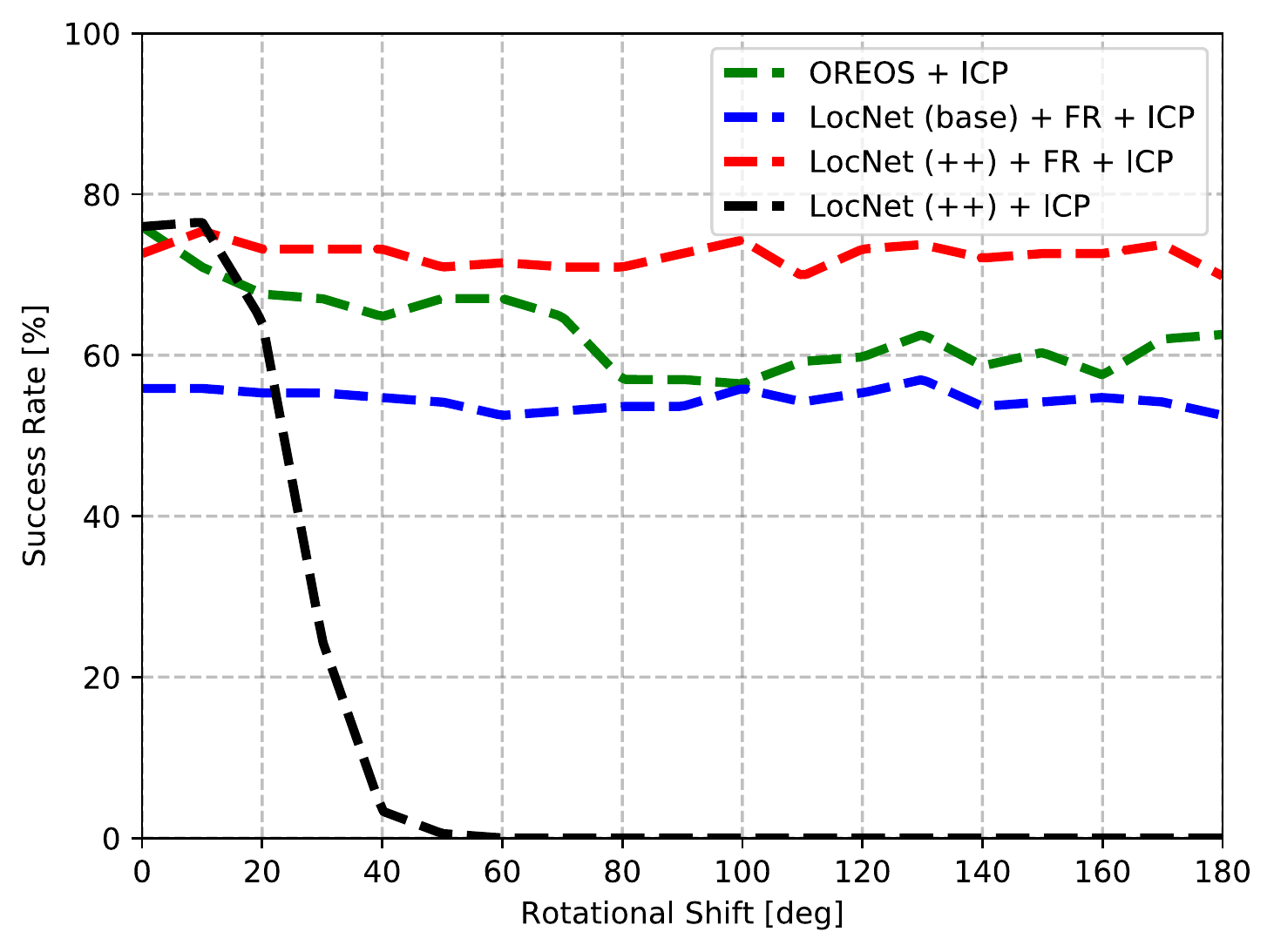}
    \label{fig:f2}
  \end{subfigure}
  \caption{\label{fig:exp:subsec:system_comp_nclt}
Metric global localization performance of \oreos on the NCLT (top) and KITTI (bottom) datasets. 
%
We compare our approach to LocNet (base) and Locnet(++), whereas the latter uses Fast Point Feature Histograms (FPFH) and RANSAC (abbreviated with FR) for the initial orientation estimation.
We vary the rotational shift of the query point cloud in $10$ degree steps in order to evaluate the orientation estimation success rate. 
%
}
\vspace{-5mm}
\end{figure}
\subsection{Metric Global Localization Performance}
\label{sec:eval:system_performance}
We evaluate the localization recall of \oreos with the recall attained by the two versions of LocNet combined with FPFH and RANSAC, for increasing discrepancies in the yaw angle between the query point cloud and the point cloud of the nearest place in the map.
For this, the query point clouds are rotated along the yaw-axis in $10$ $deg$ steps.
\begin{table*}[htp]
\begin{center}
\begin{tabular}{|c|c|c|c|c|c|c|c|c|}
\hline
\textbf{Approach}& \thead{\textbf{Preprocessing}\\$\text{[ms]}$}& \thead{\textbf{Feature Extraction}\\$\text{[ms]}$}& \thead{\textbf{CNN}\\$\text{[ms]}$}& \thead{\textbf{NN Loc}\\$\text{[ms]}$}& \thead{\textbf{FCN}\\$\text{[ms]}$}& \thead{\textbf{RANSAC}\\$\text{[ms]}$}& \thead{\textbf{ICP}\\$\text{[ms]}$}&
\thead{\textbf{Total}\\$\text{[ms]}$}\\
\hline
FPFH & - & 414 & - & - & - & 3149 & 27 & 3590\\
\hline
LocNet (base/++) & 56.5 & - & 1.0 & 1.0 & - & - & - & 58.5\\
\hline
OREOS & 12 & - & 2.37 & 1.0 & 1.0 & - & 25 & 41.37\\
\hline
\end{tabular}
\end{center}
\end{table*}
\begin{table*}[htp]
\begin{center}
\begin{tabular}{|c|c|c|c|c|c|c|c|c|}
\hline
\textbf{Approach}& \thead{\textbf{Preprocessing}\\$\text{[ms]}$}& \thead{\textbf{Feature Extraction}\\$\text{[ms]}$}& \thead{\textbf{CNN}\\$\text{[ms]}$}& \thead{\textbf{NN Loc}\\$\text{[ms]}$}& \thead{\textbf{FCN}\\$\text{[ms]}$}& \thead{\textbf{RANSAC}\\$\text{[ms]}$}& \thead{\textbf{ICP}\\$\text{[ms]}$}&
\thead{\textbf{Total}\\$\text{[ms]}$}\\
\hline
FPFH & - & 564 & - & - & - & 2124 & 24 & 2712\\
\hline
LocNet (base/++) & 79.4 & - & 1.0 & 1.0 & - & - & - & 81\\
\hline
OREOS & 19 & - & 2.89 & 1.0 & 1.0 & - & 15 & 39\\
\hline
\end{tabular}
\end{center}
\caption{\label{table:runtimes}Average computational execution times (NCLT (top) and KITTI (bottom)) in [ms].}
\end{table*}
The localization of a query point cloud is considered successful, if the following two criteria are met:
a) The nearest place candidate retrieved from the map lies within $1.5$ $m$ of the ground-truth query pose.
b) After running ICP, the refined yaw angle $\orient$ is within $2.5$ $deg$ of the ground-truth yaw angle.
Note that in this evaluation, $k=1$, that is, only the first place candidate from the map is retrieved and processed.

On NCLT it can be observed that for small discrepancy in yaw angles, \oreos and LocNet++ perform similarly, achieving approximately $60\%$ localization recall, while the original LocNet implementation performs significantly worse.
For increasing yaw discrepancies, only \oreos is able to maintain a high localization recall, demonstrating its ability to both predict accurate nearest places in the map, and estimate the yaw angle discrepancies between the query and map point clouds.
As expected, LocNet without an additional yaw estimation fails for increased yaw discrepancies, while using FPFH and RANSAC are able to achieve a localization recall between $20\%-40\%$ for misaligned point clouds.
Towards $180$ $deg$ there is an increase of the success rate of ICP for some of the methods.
This is due to the fact that in some NCLT datasets, the campus is traversed in the opposite direction.
Augmenting point clouds from these datasets by $180$ $deg$ thus results in the point clouds being already well-aligned with the map point cloud, without the need of a yaw discrepancy estimation. 
%
In addition to a decreased localization recall for large yaw angle discrepancies, the runtime of LocNet combined with FPHF and RANSAC is significantly higher than for \oreosa, as can be seen in Table~\ref{table:runtimes}.
On the KITTI dataset, the localization recall of all methods is in general higher than in case of NCLT.
This is due to the fact that the KITTI scenario is considerably simpler, with very similar driving trajectories, and without any significant environmental change.
While \oreos still performs better than LocNet (base), in this case LocNet(++) takes the lead in overall performance. 
As the in depth-analysis in Section \ref{sec:eval:pr_indepth} and Section \ref{sec:eval:yaw_indepth} will later reveal, this performance gain is mostly due to FPFH/RANSAC which almost reaches a recall of 100 $\%$. \oreos on the other hand is significantly better with the predicted orientation estimation as compared to FPFH and RANSAC, and computationally more efficient as depicted in Table~\ref{table:runtimes} and Table~\ref{table:results_orientation}. 
All approaches are evaluated on a GTX 980 Ti and an i7-4810MQ CPU @ 2.80GHz.
Preprocessing the 3D pointcloud to a 2D range image and LocNet's histograms are computed single threaded, while FPFH is implemented using PCL`s multithreaded OPM version.
%
%
%
%
%
\subsection{Place Recognition Analysis}
\label{sec:eval:pr_indepth}
\begin{figure}[H]
\centering
\includegraphics[width=0.4\textwidth]{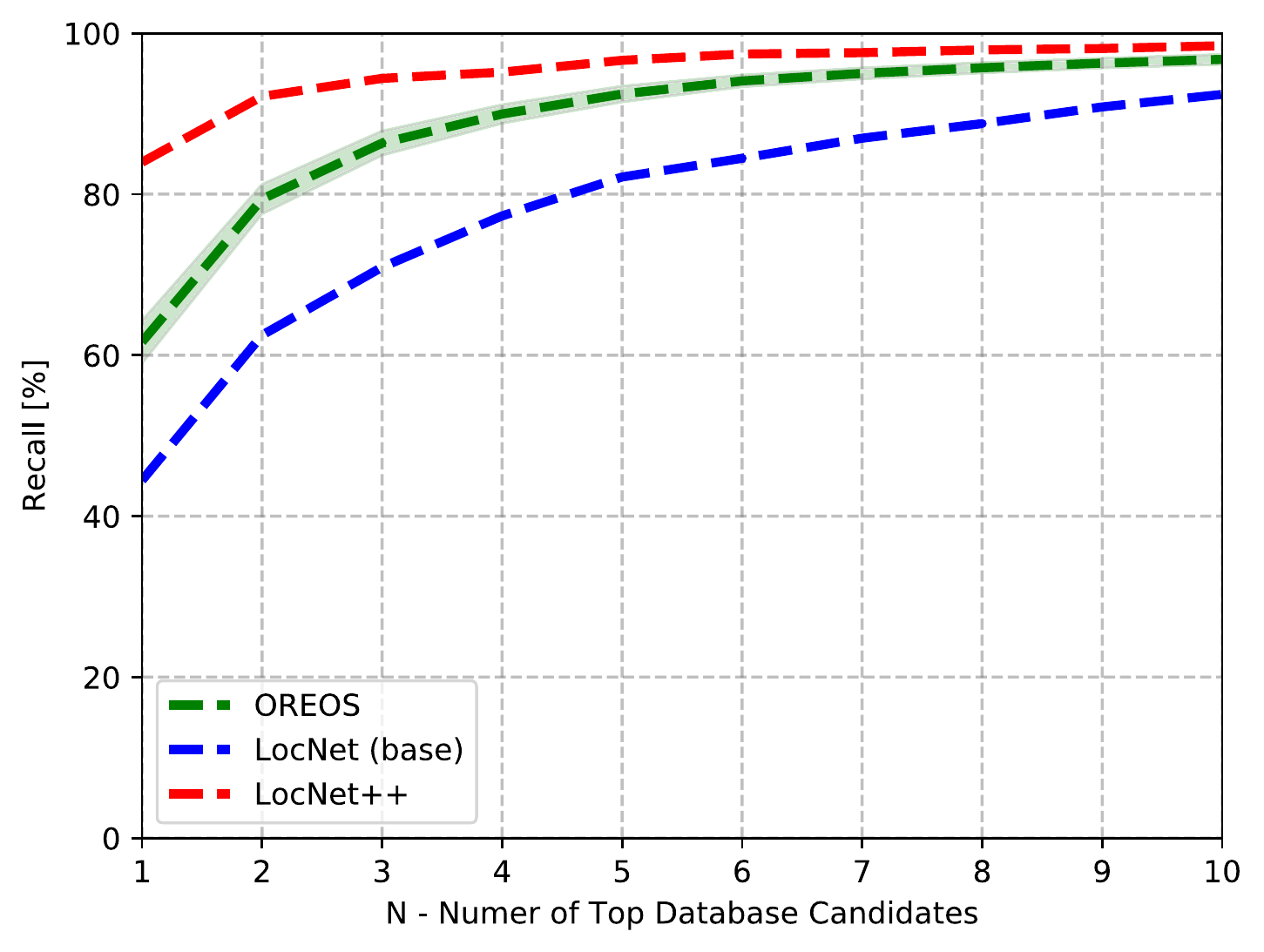}
\includegraphics[width=0.4\textwidth]{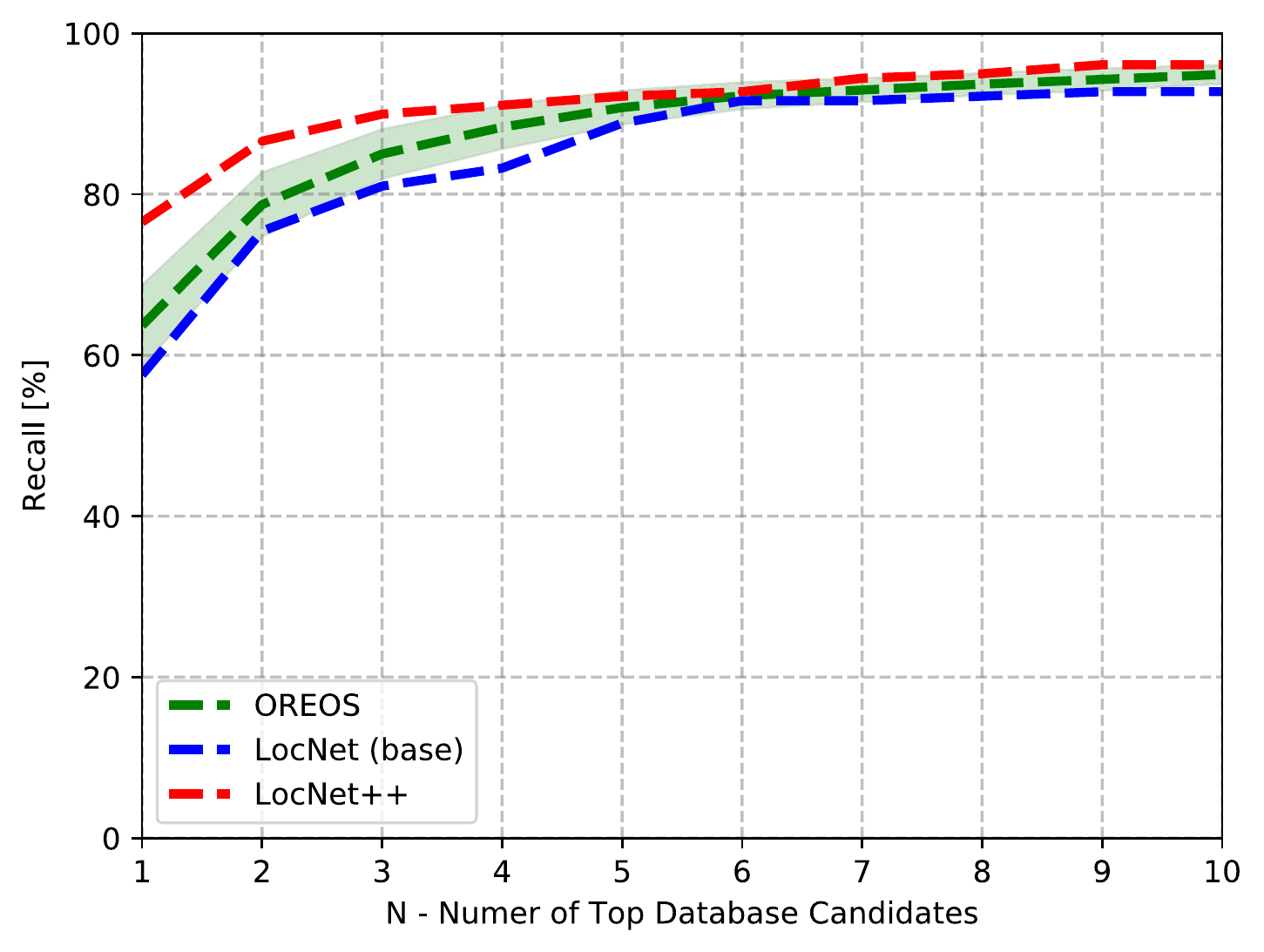}
\caption{
Place recognition performance on NCLT (top) and KITTI (bottom) of \oreosa, and the two variations of LocNet, for an increasing number of nearest place candidates retrieved from the map.
For our approach, the standard deviation over augmented rotated point clouds is shown in shaded green.
%
}
\label{fig:exp:subsec:pr_comp_nclt}
\end{figure}

%
In a practical application, it may be possible to test more than one place recognition candidate retrieved from the kd-tree.
In this section, we thus analyze the performance of the \oreos place recognition module in comparison with LocNet, for increasing values of $k$.
The respective localization recall results are shown in Figure~\ref{fig:exp:subsec:pr_comp_nclt}.
%
\oreos outperforms the LocNet base model, and attains similar performance as LocNet++ for higher values of $k$.
However, for small values of $k$, the rotation invariant histogram representation of LocNet, together with a model trained using hard negative mining, appears to exhibit an edge over the our place recognition module learned directly from the 2D range images.
As shown in Section~\ref{sec:eval:system_performance}, using the 2D range images does, however, has the advantage of allowing to also estimate a yaw angle discrepancy.
%
%

%
%
\subsection{Yaw Estimation Analysis}
\label{sec:eval:yaw_indepth}
To investigate the accuracy of the \oreos yaw angle estimation, we analyze and compare the yaw angle discrepancy estimates of our Yaw Estimation network, with the estimates generated by FPFH in combination RANSAC.
Using the ground-truth orientations of the point clouds, we can assess the estimation errors, and the respective mean and standard deviations are listed in Table~\ref{table:results_orientation}.
Both \oreos and FPFH with RANSAC exhibit similar yaw discrepancy estimation accuracy.
However, \oreos per design attains $100\%$ recall, while RANSAC is prone to fail to provide a yaw estimate in many cases.
%

\begin{table}[H]
\begin{center}
\begin{tabular}{|c|c|c|c|}
\hline
\textbf{Approach in NCLT} & \textbf{Mean} [deg] & \textbf{Std} [deg] & \textbf{Recall} [$\%$]\\
\hline
FPFH + RANSAC & 9.47 & 26.65 & 58.0 \\
\hline
OREOS & 15.95 & 21.31 & 100.0 \\
\hline
\end{tabular}
\end{center}
\begin{center}
\begin{tabular}{|c|c|c|c|}
\hline
\textbf{Approach in KITTI} & \textbf{Mean} [deg] & \textbf{Std} [deg] & \textbf{Recall} [$\%$]\\
\hline
FPFH + RANSAC & 13.28 & 32.19 & 97.0 \\
\hline
OREOS & 12.67 & 15.23 & 100.0 \\
\hline
\end{tabular}
\end{center}
\caption{\label{table:results_orientation} Absolute orientation estimation errors without ICP (NCLT (top) and KITTI (bottom)) with mean and standard deviation in degrees, and recall in $\%$.}
\end{table}

As seen in Table \ref{table:results_orientation} our approach shows a better standard deviation in degree than FPFH + RANSAC while yielding a higher recall.

%% file: conclusions.tex
\section{conclusions}
\label{sec:conclusions}
We have presented a data-driven descriptor that can be used to both retrieve near-by place candidates from a map, and estimate the yaw angle discrepancy between 3D LiDAR scans in challenging outdoor environments.
A deep Neural Network architecture is employed to learn a mapping from a range image encoding of the 3D point cloud onto a feature vector representation, which effectively encodes place and orientation dependent cues.
Using our learning approach consisting of a triplet loss approach, hard negative mining, we obtain a novel descriptor which resulting 3 DoF pose estimates set a new state-of-the-art in metric global localization for outdoor environments using only single 3D LiDAR scans. 
At the same time, our learned descriptor mapping function can be computed efficiently in real-time without discarding any useful information through handcrafted intermediate representations. 
An extensive analysis of the performance of our proposal in two different outdoor environments and sensor setups has revealed a high robustness on the orientation estimates and high place recognition recall.